\documentclass{article}
\usepackage{spconf,amsmath,graphicx,hyperref,enumitem,booktabs,multirow,adjustbox,array,tabularx}
\usepackage{bm}
\usepackage{amssymb}

\newcolumntype{Y}{>{\centering\arraybackslash}X}
\newcolumntype{Z}[1]{>{\centering\arraybackslash}m{#1}}
\title{ODSS-Net: Orthogonally Decoupled Shared-Specific Representations \\for Multi-Modal Object Re-Identification}
\name{
Zhifeng Ma$^{1}$ \quad
Yuhong Feng$^{1*}$ \quad
Chongyi Huang$^{1}$ \quad
Mingyang Zhou$^{1}$ \quad
Minwen Deng$^{2}$
\thanks{$^{*}$Corresponding author.}}
\address{$^{1}$College of Computer Science and Software Engineering, Shenzhen University, Shenzhen, China\\
$^{2}$Tencent, Shenzhen, China\\}
\begin{document}

%
\maketitle
\begin{abstract}

Multi-modal object re-identification (ReID) leverages complementary information from heterogeneous sensors. Cross-modal alignment and fusion remain challenging. Existing methods mainly reduce modality gaps via feature alignment and fusion. However, alignment may entangle modality-shared structural information with modality-specific cues, while conventional fusion may not fully exploit complementary information. To address these issues, we propose {\em ODSS-Net}, a framework for orthogonally decoupled shared-specific representation learning and adaptive fusion of complementary modality-specific features. First, a Shared Structural Feature Construction (SSFC) module suppresses modality-dependent appearance variations and extracts modality-shared structural information. Then, an Orthogonal Specific Feature Learning (OSFL) module aggregates identity-relevant local information and removes the shared-aligned component via geometric orthogonal projection. Finally, a Sample Adaptive Mixture of Experts (SAMoE) module uses the shared structure as routing context to adaptively refine and fuse modality-specific features. Experiments on three benchmarks demonstrate the effectiveness of ODSS-Net for multi-modal object ReID.
\end{abstract}
\section{Introduction}
\label{sec:intro}

Object Re-Identification ({\em ReID}) aims to retrieve specific targets across non-overlapping camera views. 
Conventional object ReID primarily relies on RGB appearance information~\mbox{\cite{he2021transreid,Dou_2023_ICCV,Li_2024_CVPR}}, which can become unreliable under illumination variations and low-light conditions.
Multi-modal object ReID mitigates these limitations by exploiting complementary information from RGB, near-infrared (NIR), and thermal infrared (TIR) modalities~\mbox{\cite{wang2024topreid,UGG-ReID2025,wang2025idea}}.
Despite their complementarity, these modalities exhibit substantial appearance discrepancies, and the reliability of modality-specific cues varies across samples.
Therefore, the key challenges lie in disentangling modality-shared and modality-specific representations and achieving sample-adaptive cross-modal fusion.

\begin{figure}[!t]
    \centering
    \includegraphics[width=\columnwidth]{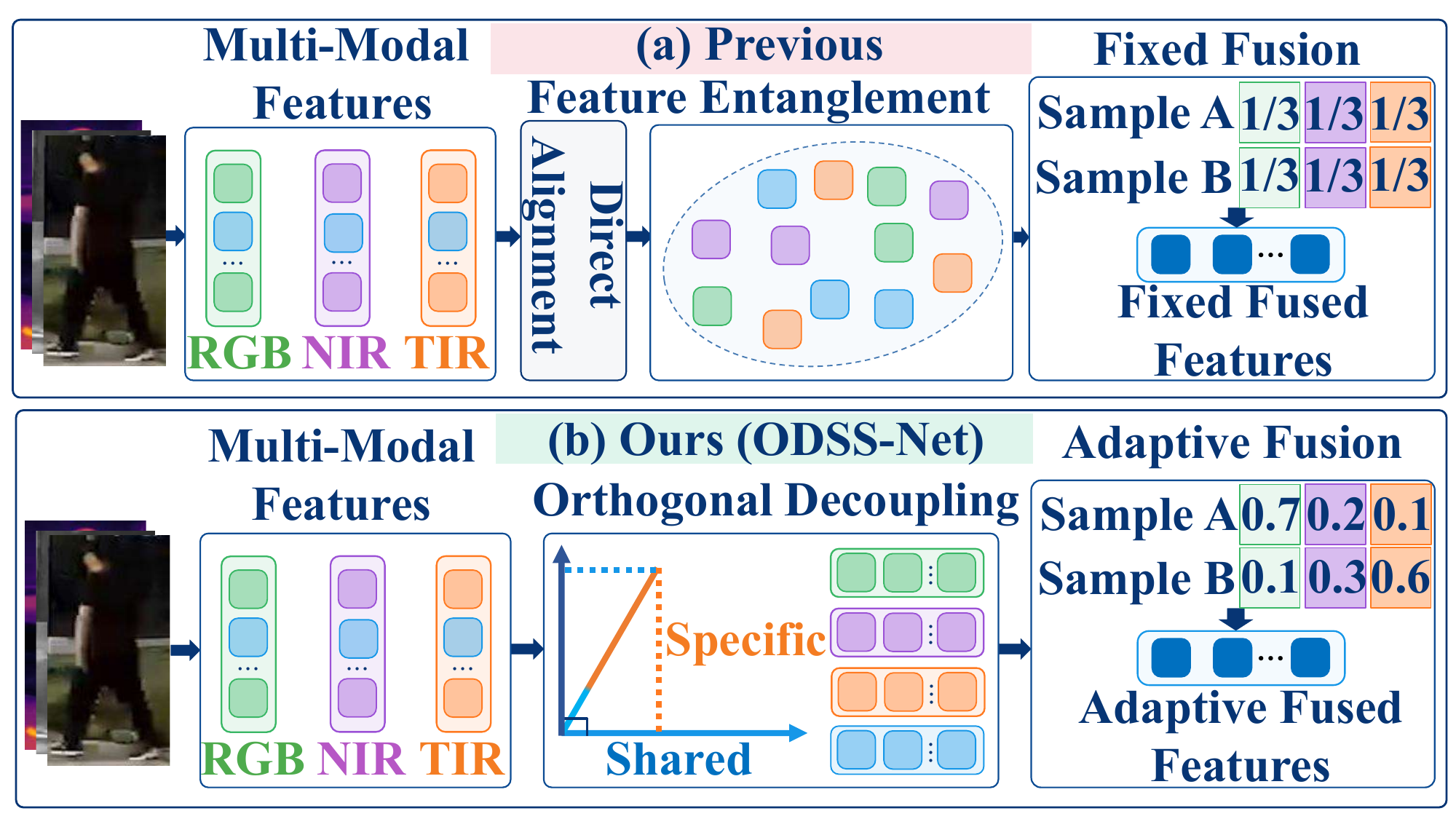}
    \caption{Motivation of ODSS-Net. (a) Feature alignment may cause shared-specific entanglement, while fixed fusion overlooks sample-wise variations. (b) ODSS-Net tackles both issues via orthogonal decoupling and sample-adaptive fusion.}
    \label{fig:motivation}
\end{figure}

\begin{figure*}[!t]
    \centering
    \includegraphics[width=\textwidth]
    {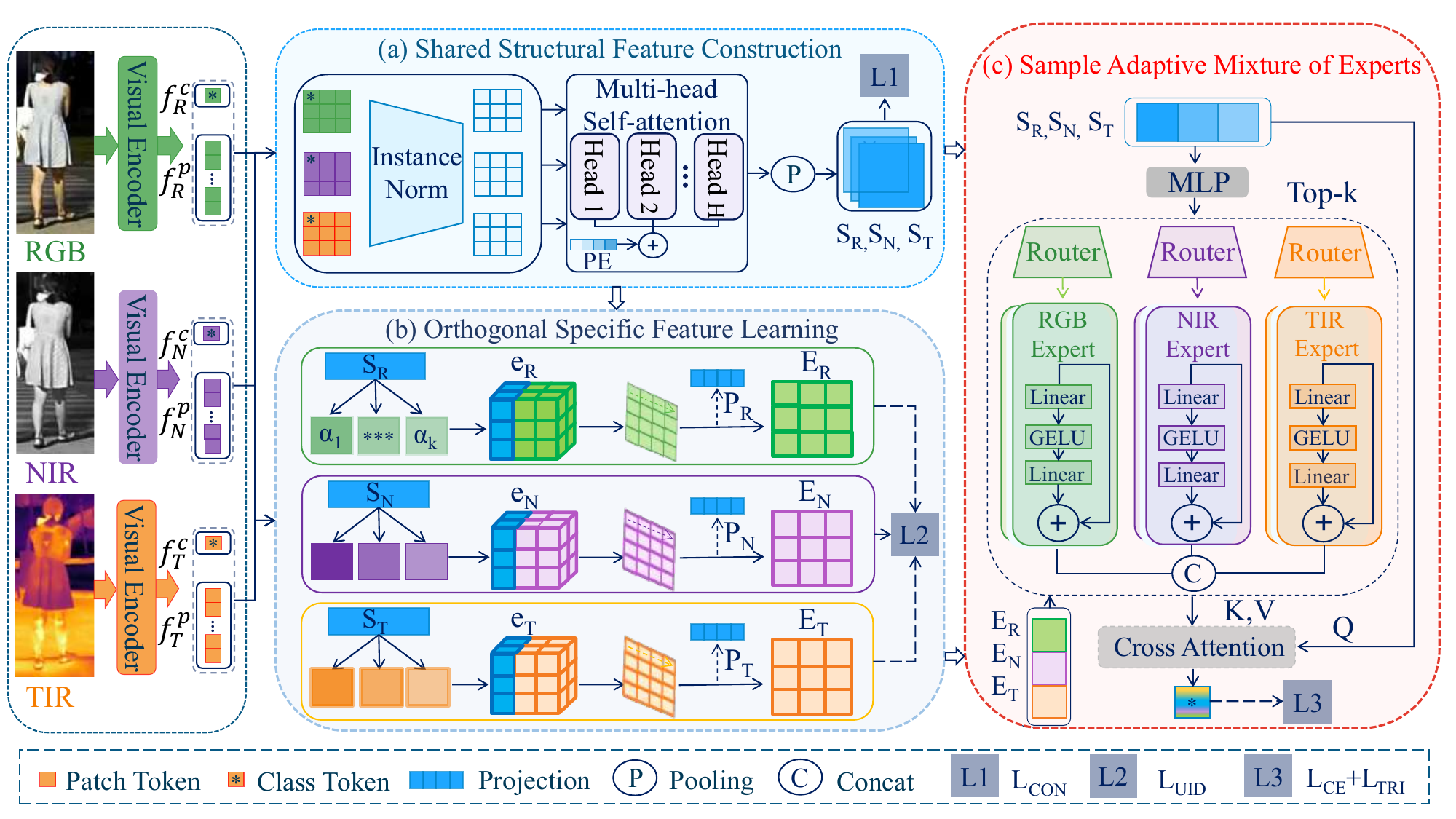}
    \caption{Overview of the proposed ODSS-Net framework. SSFC captures shared structural information by suppressing modality-dependent appearance variations. OSFL extracts identity-relevant modality-specific features via shared-guided aggregation and geometric orthogonal projection. SAMoE then adaptively refines and fuses these features under shared guidance.}
    \label{fig:framework}
\end{figure*}

As illustrated in Fig.~\ref{fig:motivation} (a), existing methods commonly reduce cross-modal discrepancies through feature alignment~\cite{wang2024topreid,liu2026signal} and cross-modal fusion~\cite{wang2025idea,wang2025mambapro}.
While direct alignment promotes modality consistency, it may also entangle shared and specific information within each modality representation. Similarly, fixed fusion applies a uniform fusion strategy to all samples, overlooking sample-wise differences in modality reliability.
Recent methods attempt to learn separate feature components~\cite{MD-ReID2025,wang2025demo}. 
However, such separation is mainly encouraged by architectural branches or training objectives, rather than being explicitly enforced during forward feature transformation.
These limitations motivate explicit shared-specific decoupling and sample-adaptive fusion.

As illustrated in Fig.~\ref{fig:motivation} (b), we propose {\em ODSS-Net}, an orthogonally decoupled shared-specific representation framework for multi-modal object ReID.
First, the Shared Structural Feature Construction ({\em SSFC}) module suppresses modality-dependent appearance variations to capture shared structural information.
Then, the Orthogonal Specific Feature Learning ({\em OSFL}) module aggregates identity-relevant local information and removes the shared-aligned component via orthogonal projection.
Finally, the Sample Adaptive Mixture of Experts ({\em SAMoE}) module leverages shared structural context to adaptively refine modality-specific features before fusion.
The main contributions are summarized as follows:

\begin{itemize}
[
    labelindent=0pt,
    leftmargin=*,
    topsep=2pt,
    partopsep=0pt,
    itemsep=2pt,
    parsep=0pt
]
    \item We propose ODSS-Net, an orthogonally decoupled shared-specific representation framework that disentangles shared and specific features and enables sample-adaptive fusion of complementary cross-modal information. Experiments on three benchmarks validate its effectiveness.
    \item We design the SSFC and OSFL modules to construct shared structural representations and explicitly decouple them from modality-specific cues via orthogonal projection.
    \item We introduce the SAMoE module to adaptively refine and fuse complementary modality-specific features.
\end{itemize}

\section{METHODOLOGY}
\label{sec:format}


\subsection{Shared Structural Feature Construction (SSFC)} \label{subsecSSFC:}

Modality-dependent appearance statistics hinder cross-modal consistency
learning in multi-modal object ReID. To address this, we introduce SSFC to suppress such variations and capture shared structural features.
Given an image \(I_m\) from modality \(m\in\mathcal{M}=\{R,N,T\}\), a weight-shared visual encoder extracts a class token \(\bm f_m^c\in\mathbb{R}^{1\times C}\) and patch tokens \(\bm f_m^p\in\mathbb{R}^{L\times C}\), which represent global semantics and local regions, respectively. 
These tokens are concatenated to form the complete token sequence:
\begin{equation}
\begin{gathered}
(\bm f_m^c,\bm f_m^p)=\mathcal{V}(I_m), \ \ \ \ 
\bm F_m=[\bm f_m^c;\bm f_m^p]
\in\mathbb{R}^{(L+1)\times C}.
\end{gathered}
\label{eq:feature_extraction}
\end{equation}

To suppress modality-dependent appearance variations, instance normalization is applied along the token dimension.
Each channel of $\bm F_m$ is standardized by its token-wise mean and variance, with a small constant $\epsilon$ for numerical stability, yielding $\bm F_m^i$, where \(i\) denotes instance normalization.

To reinforce positional information, a learnable positional embedding \(\bm P_{pos}\) is added to $\bm F_m^i$, yielding $\widetilde{\bm F}_m^i$.
The projection matrices \(\bm W_Q\), \(\bm W_K\), and \(\bm W_V\) are then applied to generate the query, key, and value representations for multi-head self-attention.
The attention operation captures regional dependencies and outputs \(\bm H_m\):
\begin{equation}
\begin{gathered}
\widetilde{\bm F}_m^i
=
\bm F_m^{\mathrm{i}}+\bm P_{\mathrm{pos}}, \\
\bm H_m
=
\operatorname{MHSA}\!\left(
\widetilde{\bm F}_m^i\bm W_Q,
\widetilde{\bm F}_m^i\bm W_K,
\widetilde{\bm F}_m^i\bm W_V
\right).
\end{gathered}
\label{eq:spatial_attention}
\end{equation}

Finally, the contextualized tokens are aggregated via average pooling, and the resulting representation is mapped to the shared feature by a lightweight MLP $\phi(\cdot)$:
\begin{equation}
\bm S_m
=
\phi\!\left(
\operatorname{AvgPool}(\bm H_m)
\right)
\in\mathbb{R}^{C}.
\label{eq:shared_feature}
\end{equation}
where the shared feature \(\bm S_m\) is used for subsequent modality-specific feature learning and sample-adaptive fusion.

\subsection{Orthogonal Specific Feature Learning (OSFL)} \label{subsec:OSFL}

Patch tokens inherently mix shared and specific information, making direct extraction of modality-specific features difficult. To address this,
the shared structural feature \(\bm S_m\) guides local aggregation by highlighting identity-related regions that are consistent across modalities.
For each modality \(m\), \(\bm S_m\) is used to weight and aggregate the patch tokens \(\bm f_m^p\in\mathbb{R}^{L\times C}\) into an identity-relevant representation:
\begin{equation}
\bm e_m =
(\bm f_m^p)^\top
\operatorname{Softmax}\!\left(
\frac{\bm f_m^p\bm S_m}{\sqrt{C}}
\right)
\in\mathbb{R}^{C}.
\label{eq:shared_guided_aggregation}
\end{equation}

Here, \(\bm e_m\) emphasizes identity-relevant local information. However, it may still retain a residual component aligned with \(\bm S_m\), since shared-guided aggregation does not explicitly separate
shared and specific information. To remove this shared residual, we apply geometric orthogonal projection {(\em GOP)}:
\begin{equation}
\bm P_m =
\frac{\bm e_m^\top\bm S_m}
{\lVert\bm S_m\rVert_2^2}\bm S_m,
\quad
\bm E_m = \bm e_m-\bm P_m.
\label{eq:exclusive_orthogonalization}
\end{equation}

Here, $\bm P_m$ is the component of $\bm e_m$ aligned
with $\bm S_m$.
Subtracting it yields the purified modality-specific feature \(\bm E_m\), which satisfies \(\bm S_m^\top\bm E_m=0\) and is used for subsequent expert refinement.

\subsection{Sample Adaptive Mixture of Experts (SAMoE)}

Since observation conditions vary across samples, a single transformation is insufficient to refine every modality-specific feature. To this end, we introduce SAMoE to select experts based on the shared context and adaptively refine each \(\bm E_m\) before fusion.
The shared features are stacked into a joint representation and then projected by an MLP \(\psi(\cdot)\) to derive the routing context:
\begin{equation}
\bm S =
[\bm S_R;\bm S_N;\bm S_T]
\in\mathbb{R}^{3\times C},
\; 
\widehat{\bm S} =
\psi\!\left(\operatorname{Flat}(\bm S)\right)
\in\mathbb{R}^{3C}.
\label{eq:shared_context}
\end{equation}
where $\mathrm{Flat}(\cdot)$ denotes the flattening operation. For each modality $m$, a dedicated router predicts expert scores, selects the top-\(k\) experts, and computes corresponding routing weights:
\begin{align}
\bm r_m
&=
\mathcal{R}_m(\widehat{\bm S})
\in\mathbb{R}^{J},
\quad
\mathcal{I}_{m,k}
=
\operatorname{TopK}(\bm r_m,k),
\label{eq:expert_routing}\\
\beta_{m,i}
&=
\begin{cases}
\displaystyle
\frac{\exp(r_{m,i})}
{\sum_{j\in\mathcal{I}_{m,k}}\exp(r_{m,j})},
& i\in\mathcal{I}_{m,k},\\[1.5mm]
0,
& i\notin\mathcal{I}_{m,k}.
\end{cases}
\label{eq:routing_weight}
\end{align}
Only the selected experts are assigned nonzero weights. Each is an independently parameterized residual MLP, and their outputs are combined as follows:
\begin{equation}
\widehat{\bm E}_m
=
\sum_{i\in\mathcal{I}_{m,k}}
\beta_{m,i}f_{m,i}(\bm E_m),
\label{eq:expert_refinement}
\end{equation}
where $f_{m,i}(\cdot)$ denotes the $i$-th expert for modality
$m$. 
The refined specific features are then stacked as follows:
\begin{equation}
\widehat{\bm E}
=
[\widehat{\bm E}_R;
 \widehat{\bm E}_N;
 \widehat{\bm E}_T]
\in\mathbb{R}^{3\times C}.
\label{eq:exclusive_concatenation}
\end{equation}

To exploit complementary specific information, $\bm S$ is used as the query and $\widehat{\bm E}$ as the key and value for cross-attention. 
The attended output is added to $\bm S$ and normalized, then concatenated with $\widehat{\bm E}$ and projected to produce $\bm U \in \mathbb{R}^{6C}$:
\begin{equation}
\bm U =
\rho\!\left(
\operatorname{Flat}\!\left(
\left[
\operatorname{LN}\!\left(
\bm S +
\operatorname{MHA}\!\left(
\bm S,\widehat{\bm E},\widehat{\bm E}
\right)
\right)
\Vert \widehat{\bm E}
\right]
\right)
\right).
\label{eq:retrieval_embedding}
\end{equation}
Here, $\Vert$ denotes feature-wise concatenation, and $\rho(\cdot)$ is the final projection that produces the retrieval embedding. 

\subsection{Objective Functions}
To align the shared features across modalities, the consistency loss is defined as follows:
\begin{equation}
\mathcal{L}_{\mathrm{CON}}
=
\frac{1}{|\mathcal{P}|C}
\sum_{(a,b)\in\mathcal{P}}
\left\lVert
\bm S_a-\bm S_b
\right\rVert_2^2,
\label{eq:consistency_loss}
\end{equation}
where $\mathcal{P}=\{(R,N),(R,T),(N,T)\}$ denotes the set of modality pairs and $C$ is the feature dimension.
This loss promotes cross-modal consistency by minimizing the pairwise differences between the shared features.

To preserve identity-discriminative cues in the orthogonalized specific features, a modality-shared identity classifier is applied, and the loss is defined as follows:
\begin{equation}
\mathcal{L}_{\mathrm{UID}}
=
-\frac{1}{|\mathcal{M}|}
\sum_{m\in\mathcal{M}}
\bm y^\top
\log
\operatorname{Softmax}
\left(\bm W_u\bm E_m\right),
\label{eq:unique_identity_loss}
\end{equation}
where $\bm y$ is the one-hot identity label, and $\bm W_u$ denotes the modality-shared identity classifier.
 
Finally, the cross-entropy loss $\mathcal{L}_{\mathrm{CE}}$~\cite{7780677} and the triplet loss $\mathcal{L}_{\mathrm{TRI}}$~\cite{hermans2017defense} are employed to supervise the entire network. 
The overall objective is defined as:

\begin{equation}
\mathcal{L}
=
\mathcal{L}_{\mathrm{CE}}
+
\mathcal{L}_{\mathrm{TRI}}
+
\lambda_1\mathcal{L}_{\mathrm{CON}}
+
\lambda_2\mathcal{L}_{\mathrm{UID}},
\label{eq:overall_objective}
\end{equation}
where $\lambda_1$ and $\lambda_2$ balance \(\mathcal{L}_{\mathrm{CON}}\) and \(\mathcal{L}_{\mathrm{UID}}\) respectively.

\section{Experiment}

\begin{table*}[t]
\centering
\caption{Performance comparison with state-of-the-art methods on three multi-modal
ReID benchmarks.}
\label{tab:sota_all}
\small
\setlength{\tabcolsep}{2pt}
\renewcommand{\arraystretch}{1}
\setlength{\aboverulesep}{0.5pt}
\setlength{\belowrulesep}{0.5pt}

\begin{tabularx}{\textwidth}{
@{}
Z{2.35cm} |
Z{1.55cm} |
Z{1.25cm} |
*{4}{Y} |
*{2}{Y} |
*{2}{Y}
@{}}
\toprule
\multirow[c]{2}{*}{\textbf{Methods}}
& \multirow[c]{2}{*}{\textbf{Publication}}
& \multirow[c]{2}{*}{\textbf{Structure}}
& \multicolumn{4}{c|}{
    \raisebox{-0.15ex}{\textbf{RGBNT201}}}
& \multicolumn{2}{c|}{
    \raisebox{-0.15ex}{\textbf{RGBNT100}}}
& \multicolumn{2}{c}{
    \raisebox{-0.15ex}{\textbf{MSVR310}}} \\
\cline{4-7}
\cline{8-9}
\cline{10-11}
& & &
\raisebox{-0.35ex}{\textbf{mAP}}
& \raisebox{-0.35ex}{\textbf{R-1}}
& \raisebox{-0.35ex}{\textbf{R-5}}
& \raisebox{-0.35ex}{\textbf{R-10}}
& \raisebox{-0.35ex}{\textbf{mAP}}
& \raisebox{-0.35ex}{\textbf{R-1}}
& \raisebox{-0.35ex}{\textbf{mAP}}
& \raisebox{-0.35ex}{\textbf{R-1}} \\
\midrule

PFNet~\cite{Zheng2021RobustMP}
& AAAI21
& CNN
& 38.5 & 38.9 & 52.0 & 58.4
& 68.1 & 94.1 & 23.5 & 37.4 \\

IEEE~\cite{Wang2022InteractEA}
& AAAI22
& CNN
& 49.5 & 48.4 & 59.1 & 65.6
& 61.3 & 87.8 & 21.0 & 41.0 \\

LRMM~\cite{wu2025lrmm}
& ESWA25
& CNN
& 52.3 & 53.4 & 64.6 & 73.2
& 78.6 & 96.7 & 36.7 & 49.7 \\

RSCNet~\cite{10772090}
& TCSVT25
& ViT
& 68.2 & 72.5 & -- & --
& 82.3 & 96.6 & 39.5 & 49.6 \\

HTT~\cite{Wang2024HeterogeneousTT}
& AAAI24
& ViT
& 71.1 & 73.4 & 83.1 & 87.3
& 75.7 & 92.6 & -- & -- \\

TOP-ReID~\cite{wang2024topreid}
& AAAI24
& ViT
& 72.3 & 76.6 & 84.7 & 89.4
& 81.2 & 96.4 & 35.9 & 44.6 \\

MambaPro~\cite{wang2025mambapro}
& AAAI25
& CLIP
& 78.9 & 83.4 & 89.8 & 91.9
& 83.9 & 94.7 & 47.0 & 56.5 \\

DeMo~\cite{wang2025demo}
& AAAI25
& CLIP
& 79.0 & 82.3 & 88.8 & 92.0
& 86.2 & \textbf{97.6} & 49.2 & 59.8 \\

IDEA~\cite{wang2025idea}
& CVPR25
& CLIP
& 80.2 & 82.1
& \underline{90.0} & \underline{93.3}
& 87.2 & 96.5 & 47.0 & 62.4 \\

MGRNet~\cite{wan2026reliable}
& TIFS26
& CLIP
& \underline{80.5} & \underline{85.0}
& \underline{90.0} & 92.6
& \underline{87.8} & \underline{97.3}
& \underline{53.2} & \textbf{67.2} \\

MCNet~\cite{MCNet2026}
& ICASSP26
& CLIP
& -- & -- & -- & --
& 82.1 & 96.1
& 38.7 & 57.5 \\

\midrule

\textbf{ODSS-Net}
& \textbf{-}
& \textbf{CLIP}
& \textbf{82.1}
& \textbf{87.1}
& \textbf{94.3}
& \textbf{95.9}
& \textbf{88.1}
& 96.2
& \textbf{54.0}
& \underline{63.3} \\[-0.2ex]

\bottomrule
\end{tabularx}
\end{table*}

\noindent\textbf{Datasets.} We evaluate our method on three multi-modal ReID datasets: RGBNT201~\cite{Zheng2021RobustMP}, RGBNT100~\cite{li2020multi}, and MSVR310~\cite{zheng2023cross}.
RGBNT201 contains 4,787 aligned RGB, NIR, and TIR triplets from 201 identities, while RGBNT100 and MSVR310 are vehicle ReID datasets containing 17,250 and 2,087 triplets, respectively.
\begin{table}[!t]
\vspace{-7mm}
\centering
\caption{Ablation study of different modules on RGBNT201.}
\label{tab:ablation}
\small

\setlength{\tabcolsep}{4pt}
\renewcommand{\arraystretch}{0.94}
\setlength{\aboverulesep}{1pt}
\setlength{\belowrulesep}{1pt}

\begin{tabular*}{\columnwidth}{
@{\extracolsep{\fill}}
c c c c c c
@{}}
\toprule
\multirow{2}{*}{\raisebox{-0.35ex}{\textbf{Model}}}
& \multicolumn{3}{c}{\raisebox{-0.2ex}{\textbf{Modules}}}
& \multicolumn{2}{c}{\raisebox{-0.2ex}{\textbf{Metrics}}} \\
\cmidrule(lr){2-4}
\cmidrule(lr){5-6}
&
\raisebox{-0.2ex}{\textbf{SSFC}}
&
\raisebox{-0.2ex}{\textbf{OSFL}}
&
\raisebox{-0.2ex}{\textbf{SAMoE}}
&
\raisebox{-0.2ex}{\textbf{mAP}}
&
\raisebox{-0.2ex}{\textbf{R-1}} \\
\midrule
A
& $\times$
& $\times$
& $\times$
& 72.9
& 78.1 \\
B
& $\checkmark$
& $\times$
& $\times$
& 74.8
& 80.1 \\
C
& $\checkmark$
& $\checkmark$
& $\times$
& 78.1
& 81.0 \\
D
& $\checkmark$
& $\checkmark$
& $\checkmark$
& \textbf{82.1}
& \textbf{87.1} \\
\bottomrule
\end{tabular*}
\end{table}

\begin{table}[t]
\vspace{-4mm}
\centering
\caption{Comparison with and without GOP on RGBNT201.}
\label{tab:gop_ablation}
\small
\setlength{\tabcolsep}{0pt}
\renewcommand{\arraystretch}{0.94}
\setlength{\aboverulesep}{1pt}
\setlength{\belowrulesep}{1pt}

\begin{tabular*}{\columnwidth}{
@{\extracolsep{\fill}}lcccc@{}
}
\toprule
\raisebox{-0.2ex}{\textbf{Model}}
& \raisebox{-0.2ex}{\textbf{mAP}}
& \raisebox{-0.2ex}{\textbf{Rank-1}}
& \raisebox{-0.2ex}{\textbf{Rank-5}}
& \raisebox{-0.2ex}{\textbf{Rank-10}}  \\
\midrule
w/o GOP & 77.0 & 81.1 & 87.7 & 91.7 \\
w/ GOP
& \textbf{82.1}
& \textbf{87.1}
& \textbf{94.3}
& \textbf{95.9} \\
\bottomrule
\end{tabular*}

\vspace{-3mm}
\end{table}

\noindent\textbf{Evaluation Metrics.}
Following common practice in ReID evaluation, we report mean average precision ({\em mAP}) and cumulative matching characteristic ({\em CMC}) at Rank-$K$, where $K \in \{1,5,10\}$.

\noindent\textbf{Implementation Details.} Our model is implemented in PyTorch and trained on an NVIDIA RTX 3090 GPU, using a pre-trained CLIP ViT-B/16~\cite{radford2021learning} as the backbone. Input images are resized to $256\times128$ for RGBNT201 and $128\times256$ for RGBNT100 and MSVR310, followed by random flipping, cropping, and erasing. The batch size is set to 64 for RGBNT201 and MSVR310, with 8 and 4 images per identity, respectively. For RGBNT100, it is 128, with 16 images per identity. All models are trained for 50 epochs using the Adam optimizer, with a learning rate of $3.5\times10^{-4}$ for the proposed modules and $5\times10^{-6}$ for the visual encoder. In SAMoE, each modality uses three experts with top-2 routing.

\noindent\textbf{Comparison with State-of-the-Art Methods.} As shown in Table~\ref{tab:sota_all}, our method ranks first on RGBNT201 across all reported metrics. It achieves 82.1\% mAP and 87.1\% Rank-1 accuracy, surpassing the second-best results by 1.6 and 2.1 percentage points, respectively. On RGBNT100 and MSVR310, it obtains the best mAP scores of 88.1\% and 54.0\%, respectively, while maintaining competitive Rank-1 scores of 96.2\% and 63.3\%. These results demonstrate that shared-guided feature decoupling and sample-adaptive expert refinement improve robustness in both person and vehicle ReID scenarios.

\noindent\textbf{Ablation Study.} As shown in Table~\ref{tab:ablation}, Model A serves as the baseline, achieving 72.9\% mAP and 78.1\% Rank-1. Adding SSFC raises these to 74.8\% and 80.1\%, validating the benefit of shared structural modeling. OSFL further improves mAP and Rank-1 to 78.1\% and 81.0\%, indicating the advantage of learning complementary modality-specific features. Finally, SAMoE achieves 82.1\% mAP and 87.1\% Rank-1 through sample-adaptive expert refinement. Overall, the full model surpasses the baseline by 9.2
and 9.0 percentage points, confirming the cumulative effectiveness of the three modules. 

\noindent\textit{Effect of GOP.} Table~\ref{tab:gop_ablation} compares $\bm e_m$ without GOP and $\bm E_m$ with GOP. Incorporating GOP improves mAP and Rank-1 from 77.0\% and 81.1\% to 82.1\% and 87.1\%, and provides cleaner modality-specific cues for subsequent adaptive fusion.

\begin{figure}[t]
    \vspace{-4mm}
    \centering
    \includegraphics[width=\columnwidth]
    {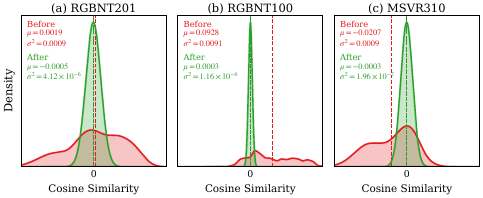}
    \caption{Cosine similarity distributions of \((S_m,e_m)\) before and \((S_m,E_m)\) after geometric orthogonal projection.}
    \label{fig:cosine_distribution}
    \vspace{-3mm}
\end{figure}

\noindent\textbf{Cosine Similarity Distributions.} As shown in Fig.~\ref{fig:cosine_distribution}, GOP sharply concentrates the similarities around zero when transforming \((\bm S_m,\bm e_m)\) to \((\bm S_m,\bm E_m)\) across all benchmarks. This concentration indicates reduced shared-specific correlation and more effective orthogonal decoupling.

\section{CONCLUSION}
We propose ODSS-Net, a framework for orthogonally decoupled shared-specific representation in multi-modal object ReID. 
Specifically, SSFC constructs modality-shared structural representations, OSFL separates modality-specific cues via geometric orthogonal projection, and SAMoE performs sample-adaptive refinement and fusion under shared guidance. 
Extensive experiments on three benchmarks demonstrate its effectiveness and highly competitive performance.



\vfill\pagebreak

\bibliographystyle{IEEEbib}
\bibliography{strings,refs}

\end{document}